\documentclass{article}
\usepackage{spconf,amsmath,graphicx}
\newcommand\Tau{\mathcal{T}}

\title{Exemplar-Free Online Continual Learning}
\name{Jiangpeng He and Fengqing Zhu}
\address{
Elmore Family School of Electrical and Computer Engineering \\ 
Purdue University, West Lafayette, Indiana, 47906, USA
}
\begin{document}
%
\maketitle
\begin{abstract}
Targeted for real world scenarios, online continual learning aims to learn new tasks from sequentially available data under the condition that each data is observed only once by the learner. Though recent works have made remarkable achievements by storing part of learned task data as exemplars for knowledge replay, the performance is greatly relied on the size of stored exemplars while the storage consumption is a significant constraint in continual learning. In addition, storing exemplars may not always be feasible for certain applications due to privacy concerns. In this work, we propose a novel exemplar-free method by leveraging nearest-class-mean (NCM) classifier where the class mean is estimated during training phase on all data seen so far through online mean update criteria. We focus on image classification task and conduct extensive experiments on benchmark datasets including CIFAR-100 and Food-1k. The results demonstrate that our method without using any exemplar outperforms state-of-the-art exemplar-based approaches with large margins under standard protocol (20 exemplars per class) and is able to achieve competitive performance even with larger exemplar size (100 exemplars per class).
\end{abstract}
\section{Introduction}
\label{sec:intro}
Though modern deep learning based approaches have achieved significant progress to address computer vision problems such as image recognition, it is still challenging to learn new tasks incrementally from data stream due to the unavailability of learned task data. The major obstacle is called catastrophic forgetting~\cite{CF} where the performance on old tasks drop dramatically during the learning phase of new task. To overcome this issue, online continual learning~\cite{ILIO,GEM,A-GEM} has emerged, which defines the learning protocol that both new tasks and their data come sequentially overtime and each data is used only once for training. During inference, the model should perform well on all tasks learned so far without knowing the task index. While existing methods~\cite{prabhu2020gdumb_online,shim2020online_ASER} have made remarkable progress by storing part of learned task data as exemplars during continual learning, there are several drawbacks associated with exemplar-based approaches: (i) it requires extra storage consumption, which is a significant constraint for online continual learning; (ii) it poses a new challenging problem of how to select the most representative data as exemplars, (iii) for certain applications such as health or medical research, the data may not be allowed to be kept for a long time due to privacy concern. In this work, we propose a novel exemplar-free online continual learning method for image classification task, which addresses the aforementioned limitations of current approaches.

\begin{figure}[t]
\begin{center}
  \includegraphics[width=.9\linewidth]{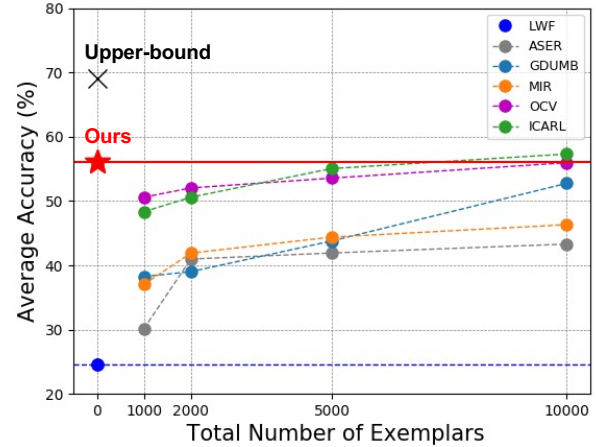}
  \vspace{-0.3cm}
  \caption{CIFAR-100 Top-1 average accuracy after learning all tasks with incremental step size 5. Dash lines show the results of existing work, which require stored exemplars (except LWF) and the red solid line shows result of our method. Upper-bound is obtained by training a model using all training samples from all classes.}
  \label{fig:intro}
\end{center}
\end{figure}

One of the main reasons for catastrophic forgetting is the biased predictions caused by biased parameters in the classifier towards new classes due to the lack of old data~\cite{BIC}. The most recent work~\cite{He_2022_WACV} addresses this problem by selecting candidates at first and then performing classification using distance-based classifier~\cite{PNN} based on stored exemplars. Inspired by this, we instead leverage nearest-class-mean(NCM) classifier, which uses class mean vector for classification and does not require any exemplar. In addition, compared with the NCM used in ICARL~\cite{ICARL} where the class mean is estimated using stored exemplars, our mean vector for each seen class is calculated on all data seen so far during training phase through online mean update criteria, which is more representative especially when the allowed exemplar size is limited. Furthermore, our NCM is performed only on selected candidates~\cite{He_2022_WACV} so the class mean are better separated than using all classes as in ICARL~\cite{ICARL}, thus achieving higher accuracy for classification. 

As shown in Fig~\ref{fig:intro}, without using any exemplar, our method applied on CIFAR-100~\cite{CIFAR} not only outperforms existing methods~\cite{ICARL,prabhu2020gdumb_online,He_2022_WACV,LWF,shim2020online_ASER,NEURIPS2019_15825aee_MIR} with large margins under standard experimental protocol as proposed in~\cite{ICARL} (2,000 exemplars in total), but also achieves competitive performance given increased exemplar size. The main contributions are summarized below. 
\begin{itemize}
    \item We propose a novel exemplar-free online continual learning method by leveraging NCM classifier with class mean estimated on all data seen so far.
    \item We conduct extensive experiments on CIFAR-100~\cite{CIFAR} and Food-1k~\cite{Food2K} to show the effectiveness of our method for different datasets by varying both the exemplar size and incremental step size.
    
\end{itemize}

\section{Related Work}
\label{related work}
With the objective of mitigating catastrophic forgetting, continual learning has been studied in both offline~\cite{BIC, mainatining, rebalancing, he2021unsupervised,ICARL,EEIL,LWF} and online scenarios~\cite{ILIO,GEM,A-GEM,prabhu2020gdumb_online,NEURIPS2019_15825aee_MIR,He_2022_WACV,He_2021_ICCVW}. In the online scenarios, each data is observed only once by the model, which is more related to real life applications. In this section, we summarize existing methods that are closely related to our work. 

\textit{Regularization-based} methods retain learned knowledge by restricting the change of corresponding weights. Knowledge distillation loss~\cite{KD} is widely used in~\cite{LWF, ICARL, BIC} and a variant distillation loss is introduced in ILIO~\cite{ILIO} to achieve improved performance in online scenario. Besides, A-GEM~\cite{A-GEM} is an efficient version of GEM~\cite{GEM} where both methods use stored exemplars to ensure that the loss for learned tasks does not increase during each learning step. Most recently, OFR~\cite{He_2021_ICCVW} proposed a novel clustering based exemplar selection approach and showed its effectiveness on food image classification task. 

\textit{Reply-based} approach aims to address catastrophic forgetting by storing part of old task data to perform knowledge replay during continual learning. ICARL~\cite{ICARL} proposed to apply herding algorithm~\cite{HERDING} to select and store exemplars based on class mean. Random retrieval is applied in Experience-Replay(ER)~\cite{ER_1,ER_2} to ensure that each new data has the same probability to be stored as exemplar in memory buffer. MIR~\cite{NEURIPS2019_15825aee_MIR} proposed a controlled sampling of memories. A greedy balancing sampler was introduced in GDUMB~\cite{prabhu2020gdumb_online} which randomly selected as much data as the memory allowed and the classifier was trained on stored exemplars only. A exemplar scoring method was proposed in ASER~\cite{shim2020online_ASER} to preserve latent decision boundary. Instead of using original data as exemplar, OCV~\cite{He_2022_WACV} only selected and stored extracted feature embeddings. As mentioned in Section~\ref{sec:intro}, the use of exemplars may not always be feasible in real life. In contrast, our proposed method does not require to store any exemplar while achieving competitive performance compared to exemplar-based methods even for very large exemplar sizes.

\begin{figure}[t]
\begin{center}
  \includegraphics[width=1.\linewidth]{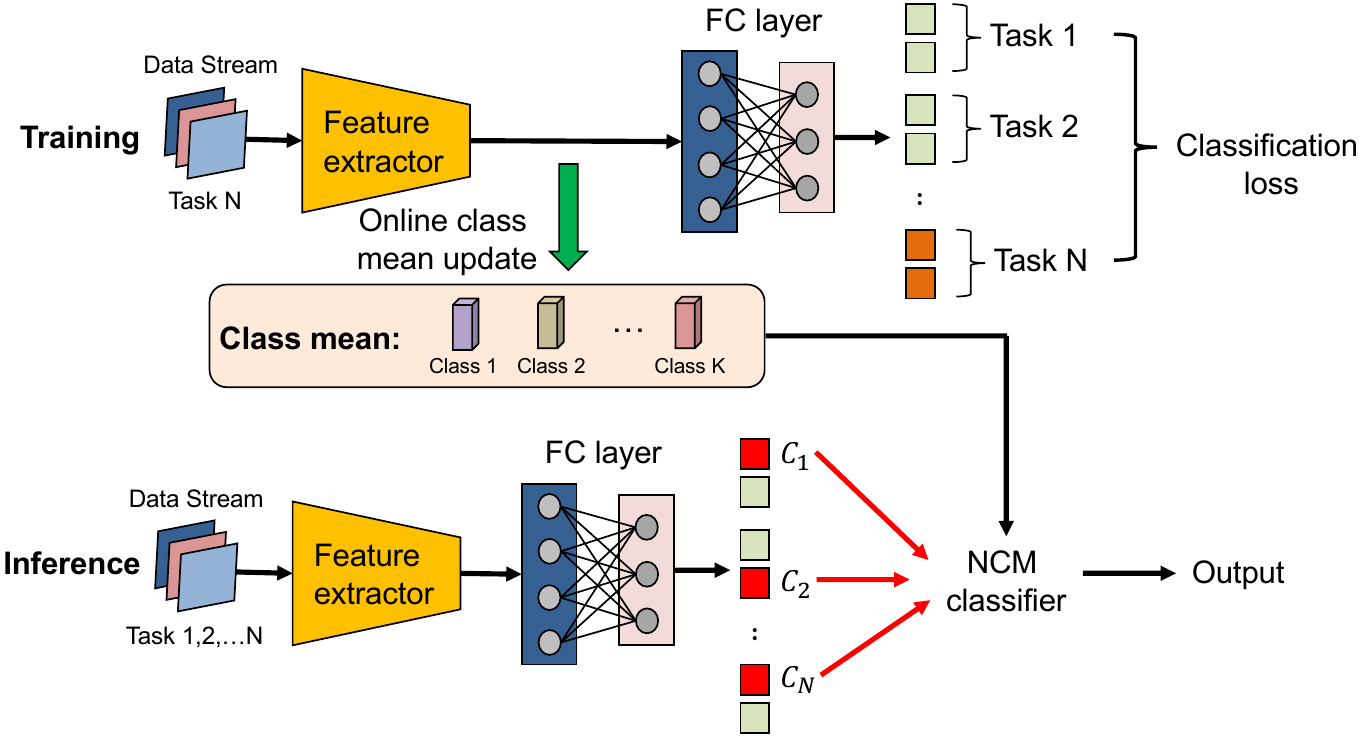}
  \vspace{-0.4cm}
  \caption{The overview of our method. The upper half shows the online training phase where we estimate the class mean dynamically using online mean update criteria on all data seen so far. The FC output for learned tasks are fixed to maintain the discrimination. For the inference phase, we first select candidates from each learned task, denoted as $C_1,...C_N$, and then apply NCM classifier for classification based on estimated class mean $Class_{C_1},...Class_{C_N}$}.
  \label{fig:method}
\end{center}
\end{figure}

\begin{figure*}[t]
\begin{center}
  \includegraphics[width=1.\linewidth]{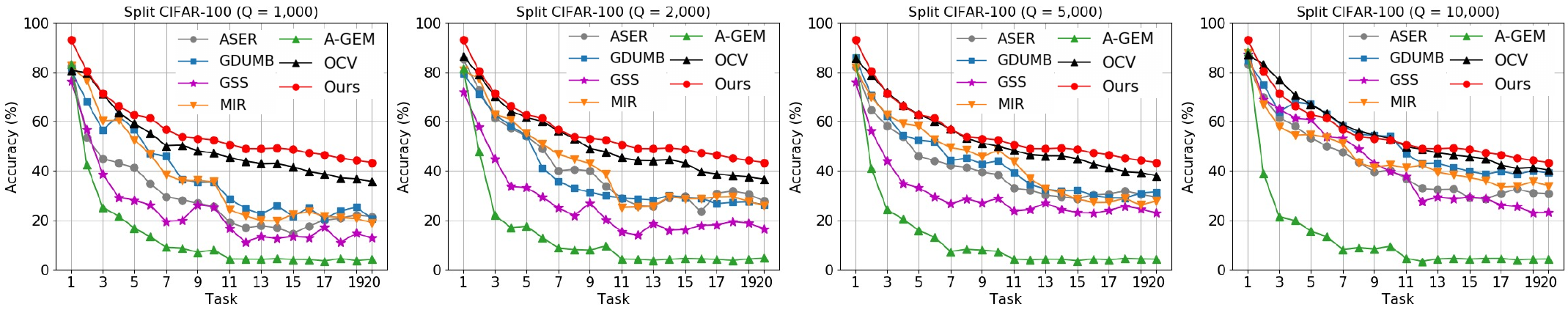}
  \vspace{-0.7cm}
    \caption{\textbf{Results on Split CIFAR-100} by varying the allowed exemplar size $Q \in \{1,000, 2,000, 5,000, 10,000\}$. There are $20$ tasks in total with each task contains $5$ non-overlapped classes.}
  \label{fig:cifari100}
\end{center}
\end{figure*}

\section{Method}
\label{sec:method}
\subsection{Problem Formulation}
\label{sub:preliminary}
The continual learning for image classification problem $\Tau$ can be formulated as learning a sequence of $N$ tasks $\{\Tau^1,...,\Tau^N\}$ corresponds to $N$ incremental learning steps where each task contains $M$ non-overlapped classes, which is also known as incremental step size. Let $\{D^1, ..., D^N\}$ denote all training data, where $D^i = \{(\textbf{x}_1^i,y_1^i)...(\textbf{x}_{n_i}^i,y_{n_i}^i)\}$ indicates the training data that belongs to the added $M$ new classes of the incremental step $i$, $\textbf{x}$ and $y$ represent the data and the label respectively, and $n_i$ refers to the number of total training data in $D^i$. During the learning phase of $\Tau^N$, only $D^N$ is available to use and each data is observed only once by the model in an online learning fashion. After incremental step $N$, the updated model should perform classification task on all seen classes belonging to $\{\Tau^1,...,\Tau^N\}$.

\subsection{Proposed Method}
\label{sub:our method}
The overview of our proposed method is shown in Fig~\ref{fig:method}. A fixed feature extractor pretrained on large scale image datasets, \textit{e.g.}, ImageNet \cite{IMAGENET1000} is applied as backbone network in both training and inference phases, which provides more discriminative embeddings than original images as input for online continual learning~\cite{He_2022_WACV}.

\subsubsection{Training Phase}
\label{subsub:training}
The Nearest-Mean-of-Exemplar(NME) classifier proposed in ICARL~\cite{ICARL} achieves remarkable progress. However, the performance greatly relies on the exemplar size as the class mean vectors used for classification is only estimated through stored exemplars, which struggles when allowed storage is limited or the selected exemplars are not representative enough. As shown in the upper half of Fig~\ref{fig:method}, our method addresses this issue by estimating the class mean on all data seen so far using online mean update criteria. Specifically, during the learning phase of $\Tau^N$, for each new data $(\textbf{x}_i^N, y_i^N)$, we calculate the class mean vector using Eq~(\ref{eq:onlinemean}).
\begin{equation}
\label{eq:onlinemean}
\textbf{v}_{y_i^N} = \frac{n}{n+1} \textbf{v}_{y_i^N} + \frac{1}{n+1} \mathcal{F}(\textbf{x}_i^N)
\end{equation}
where $\textbf{v}$ denotes the class mean, which is initialized as zero for each new class. $\mathcal{F}$ refers to the feature extractor and $n$ is the number of data seen so far for class $y_i$. Instead of using knowledge distillation loss~\cite{KD} for regularization, we apply cross-entropy as classification loss to maximally maintain the discrimination for each learned task~\cite{He_2022_WACV}, which provides the basis for selecting candidates in the inference phase.

\subsubsection{Inference Phase}
\label{subsub:inference}
As introduced in Section~\ref{sec:intro}, catastrophic forgetting is largely due to the prediction bias towards new classes. A recent work~\cite{He_2022_WACV} addressed this issue by selecting candidates at first and then applying a distance-based classifier that does not have biased parameters. However, it still stores exemplars and require synthesized data to tune hyper-parameters in final prediction equation. Our method addresses this problem by leveraging nearest-class-mean (NCM) classifier on selected candidates for classification. Specifically, after the learning phase of $\Tau^N$, we denote $\{o^1_1, o^1_2,...o^1_M,...,o^N_1, o^N_2,...o^N_M\}$ as the output of the FC layer where $o^i_j$ refers to the output logit for the class $j$ in task $i$ and M is the incremental step size. The total number of classes seen so far is $K = M \times N$. For the output of each task $i\in\{1,2,...N\}$, we select candidate by $C_i = \textit{argmax}\{o^i_1,...o^i_M\}$. Finally, the NCM classifier make prediction for test data $\textbf{x}_t$ by using Eq~(\ref{eq:ncm}).
\begin{equation}
    \label{eq:ncm}
    \begin{split}
             &y_{\textbf{x}_t} = \mathop{\textit{argmin}}_{C_1,C_2,...,C_N}\{d^{C_1},...d^{C_N}\} \\
            & where \quad  d^i = ||\textbf{v}_i - \mathcal{F}(\textbf{x}_t)||_2
    \end{split}
\end{equation}
where $d^i, i\in\{C_1,...C_N\}$ denotes the Euclidean distance between class mean $\textbf{v}_i$ and extracted embedding of test data $\mathcal{F}(\textbf{x}_t)$. Compared with existing approaches~\cite{ICARL,He_2022_WACV}, our method neither requires storing exemplars nor needs to tune any hyper-parameter. 

\begin{figure*}[t]
\begin{center}
  \includegraphics[width=.9\linewidth]{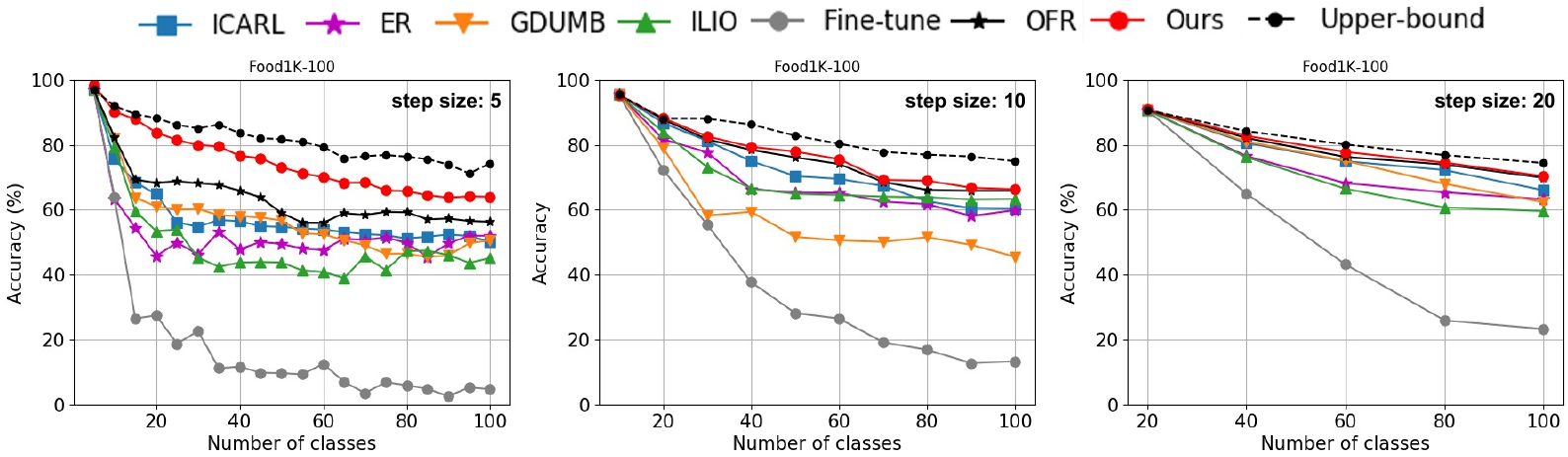}
  \vspace{-0.5cm}
    \caption{\textbf{Results on Food1k-100} by varying the incremental step size $M \in \{5, 10, 20\}$ (Best viewed in color)}
  \label{fig:food1k100}
\end{center}
\end{figure*}

\begin{table}[t!]
    \centering
    \scalebox{.72}{
    \begin{tabular}{|ccccccccc|}
        \hline
        Datasets & \multicolumn{8}{c|}{\textbf{Split CIFAR-100}} \\
        \hline
        Size of exemplar set & \multicolumn{2}{c}{$Q = 1,000$} & \multicolumn{2}{c}{$Q = 2,000$} & \multicolumn{2}{c}{$Q = 5,000$} & \multicolumn{2}{c|}{$Q = 10,000$}\\
        \hline
        Accuracy(\%) & Avg & Last & Avg & Last & Avg & Last & Avg & Last  \\
        \hline
        \hline
        A-GEM~\cite{A-GEM}&13.9 & 4.3 & 14.1 & 4.83 & 13.8 & 4.4 & 13.9 & 4.5 \\
        MIR~\cite{aljundi_mir} & 37.4 & 19.1& 41.9 & 26.1 & 44.4 & 27.9 & 46.3 & 34.0 \\
        GSS~\cite{NEURIPS2019_e562cd9c_GSS} & 24.2 & 12.9 & 26.9 & 16.7 & 31.5 & 23.1 & 43.2 & 23.3 \\
        ASER~\cite{shim2020online_ASER} & 29.9 & 21.6 & 40.9 & 27.8 & 41.2 &29.5 & 43.3 & 30.8\\
        GDUMB~\cite{prabhu2020gdumb_online} & 38.3 & 20.7 & 39.2 & 26.4 & 43.8 & 31.6 & 52.8 & 39.2\\
        OCV~\cite{He_2022_WACV} & 50.6 & 35.8 & 52.0 & 36.7 & 53.6 & 38.1 & 56.0 & 40.3\\
        \hline
        Ours (w/o) & & Avg: & 54.7 & & Last:& 41.6 & & \\
        Ours & & Avg: & \textbf{56.3} & & Last:& \textbf{43.4} & & \\
        \hline
    \end{tabular}
    }
    \caption{\textbf{Average accuracy and Last step accuracy on Split CIFAR-100}. Best results marked in bold.  }
    \label{tab:cifar}
\end{table}

\section{Experimental Results}
\label{sec:exp}
We validate our method on two benchmark datasets including CIFAR-100~\cite{CIFAR} and Food-1k~\cite{Food2K}. For CIFAR-100, we follow the protocol in~\cite{zenke2017continual_splitcifar} to construct \textbf{Split CIFAR-100} by dividing the 100 categories into 20 tasks, each contains 5 classes. For Food-1k, same as~\cite{He_2021_ICCVW}, we first randomly select 100 food categories to construct \textbf{Food1k-100} and then divide the subset into 5, 10, and 20 splits to conduct experiments. 

\subsection{Implementation Detail}
\vspace{-0.2cm}
\label{sub:implementation detail}
We follow the benchmark experimental protocol in~\cite{GEM,ICARL} to use ResNet-18~\cite{RESNET} as backbone network. We use SGD optimizer with fixed learning rate of $0.1$. Batch size is $16$ and each training data is used only once (1 epoch). The splits of both datasets uses identical random seed as in~\cite{ICARL,He_2021_ICCVW} and the backbone network for all compared methods is pre-trained on ImageNet~\cite{IMAGENET1000} to ensure fair comparison in all experiments. we use top-1 accuracy as the evaluation metric.

\vspace{-.4cm}

\subsection{Results on Split CIFAR-100}
\vspace{-.2cm}
\label{sub:split_cifar100}
We compare our method with \textbf{ASER}~\cite{shim2020online_ASER}, \textbf{GDUMB}~\cite{prabhu2020gdumb_online}, \textbf{GSS}~\cite{NEURIPS2019_e562cd9c_GSS},  \textbf{MIR}~\cite{NEURIPS2019_15825aee_MIR}, \textbf{A-GEM}~\cite{A-GEM} and OCV~\cite{He_2022_WACV}. We vary the exemplar size $Q$ for existing methods with $Q\in \{1,000, 2,000, 5,000, 10,000\}$ and the result is shown in Fig~\ref{fig:cifari100}. We observe significant improvements for the exemplar size $Q\in \{1,000, 2,000\}$ while still achieving competitive performance with larger $Q$. Note that $Q=2,000$ is a standard protocol~\cite{ICARL} for exemplar-based approaches. Although storing more exemplars will result in performance improvements for exemplar-based methods, it requires extra storage memory, which is a significant constraint for online continual learning and may not always be feasible in real life applications. The average accuracy \textit{Avg} and last step accuracy \textit{Last} are summarized in Tab~\ref{tab:cifar} where \textit{Avg} is calculated by averaging all accuracy obtained after each learning step, which shows the overall performance for the entire online continual learning problem and the \textit{Last} accuracy shows the performance after the continual learning for all classes seen so far. Our method achieves the best results in terms of \textit{Avg} and \textit{Last} while does not require storing exemplars compared with existing work. Besides, we also include \textbf{Ours(w/o)} for comparison, which performs NCM on all classes instead of on candidates as in \textbf{Ours}, the result shows the effectiveness of selecting candidates when applying NCM for classification under online continual learning scenario. 

\vspace{-0.3cm}

\subsection{Results on Food1k-100}
\vspace{-0.2cm}
\label{sub:split_food1k-100}
In this section, we evaluate our method using the challenging food images. Therefore, besides comparing with existing methods: \textbf{ICARL}~\cite{ICARL}, \textbf{ER}~\cite{ER_1,ER_2}, \textbf{GDUMB}~\cite{prabhu2020gdumb_online}, \textbf{ILIO}~\cite{ILIO} and \textbf{OFR}~\cite{He_2021_ICCVW}, we follow the experimental setting in~\cite{He_2021_ICCVW} to further include \textbf{Fine-tune} (using only new class data to update model without considering the learned task performance) as baseline and \textbf{Upper-bound} (training the model using all training samples from all seen classes at each step) for experiment. The exemplar size is fixed with $Q=2,000$ as in protocol~\cite{ICARL, He_2021_ICCVW} and we vary the step size for 5, 10 and 20 corresponding to 20, 10 and 5 incremental steps, respectively. The result is shown in Fig~\ref{fig:food1k100}. We observe severe performance degradation by comparing \textbf{Fine-tune} with \textbf{Upper-bound} due to the lack of training data for learned tasks, which shows the necessity to address catastrophic forgetting problem for online incremental learning. In addition, by comparing \textbf{Fine-tune} and \textbf{Upper-bound} results for different step sizes, we notice that the problem becomes more challenging when the step size is smaller due to the increase of incremental steps. Our proposed method not only achieves the best results with smallest performance gap between \textbf{Upper-bound} for all step sizes, but also outperforms the existing work with a larger margin in the challenging case of small step size.


\vspace{-0.3cm}
\section{Conclusion}
\vspace{-0.2cm}
\label{sec:conclusion}
In this paper, we focus on the online continual learning for image classification problem. We propose a novel exemplar-free method by leveraging nearest-class-mean (NCM) classifier based on class mean estimated on all data seen so far during the training phase through online mean update criteria. In addition, we apply NCM on selected candidates only instead of all classes to improve the performance. Compared with state-of-the-arts, our method neither requires storing exemplars nor contains hyper-parameters tuning while still achieving promising results on CIFAR-100 by varying exemplar size $Q \in \{1,000, 2,000, 5,000, 10,000\}$ for existing approaches. Besides, we validate our method on the challenging Food-1k dataset and show improved performance for different incremental step sizes $M\in\{5, 10, 20\}$. Our future work will focus on a more realistic setting for unsupervised continual learning where the class label is not provided. One possible solution is to update the model by using pseudo labels as shown in the recent work~\cite{he2021unsupervised}.

\bibliographystyle{IEEEbib}
{\small\bibliography{strings,refs}}

\end{document}